%% file: egpaper_for_review.tex
\documentclass[10pt,twocolumn,letterpaper]{article}

\usepackage{iccv}
\usepackage{times}
\usepackage{epsfig}
\usepackage{graphicx}
\usepackage{amsmath}
\usepackage{amssymb}
\usepackage{subcaption}
\usepackage{blindtext}
\usepackage{float}
\usepackage{hyperref}
\hypersetup{pagebackref=true,breaklinks=true,letterpaper=true,colorlinks,bookmarks=false}

\iccvfinalcopy % *** Uncomment this line for the final submission

\def\iccvPaperID{1979} % *** Enter the ICCV Paper ID here

\ificcvfinal\fi

\begin{document}

%%%%%%%%% TITLE
\title{LIFE: Lighting Invariant Flow Estimation}
% \title{Learning Flow Estimation in General Challenge Scenarios}

\author{
Zhaoyang Huang$^{1,3}$\thanks{}\\
Ka Chun Cheung$^{3}$
\and
Xiaokun Pan$^{2*}$\\
Guofeng Zhang$^{2}$
\and
Runsen Xu$^{2}$\\
Hongsheng Li$^{1}$
\and
Yan Xu$^{1}$
\and
$^{1}$CUHK-SenseTime Joint Laboratory, The Chinese University of Hong Kong \\
$^{2}$State Key Lab of CAD\&CG, Zhejiang University\\
$^{3}$NVIDIA AI Technology Center, NVIDIA
}

% \maketitle
\ificcvfinal\thispagestyle{empty}\fi

\twocolumn[{%
\maketitle
\begin{figure}[H]
\hsize=\textwidth
    \centering
    \vspace{-3em}
    % \resizebox{2.0\linewidth}{!}{
    \begin{subfigure}[b]{0.5\linewidth}
        \includegraphics[width=\linewidth]{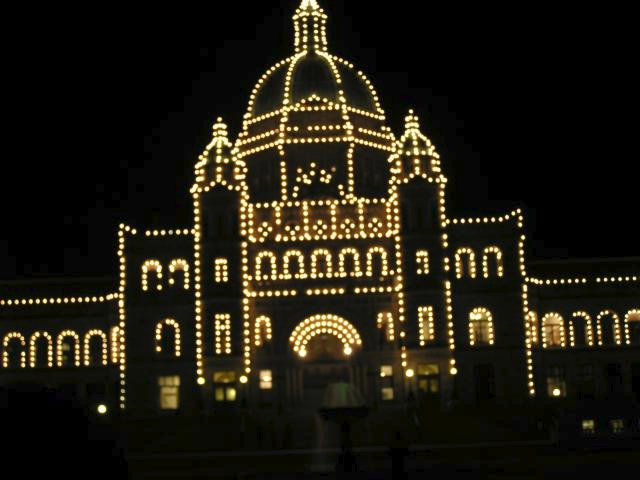}
        \caption{\small Source Image}
    \end{subfigure}
    \begin{subfigure}[b]{0.5\linewidth}
        \includegraphics[width=\linewidth]{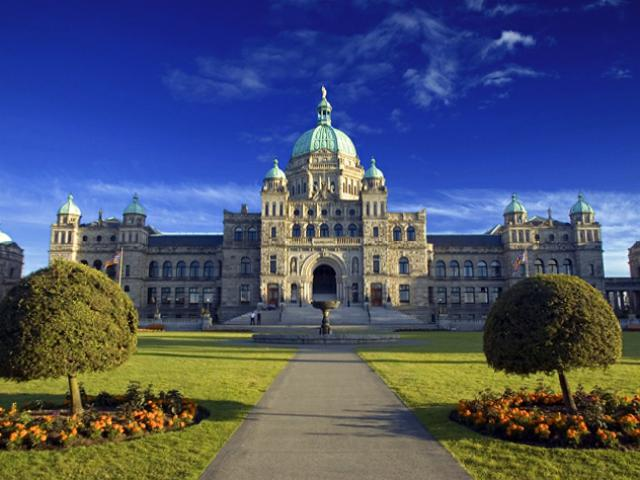}
        \caption{\small Target Image}
    \end{subfigure}
    \begin{subfigure}[b]{0.5\linewidth}
    \includegraphics[width=\linewidth]{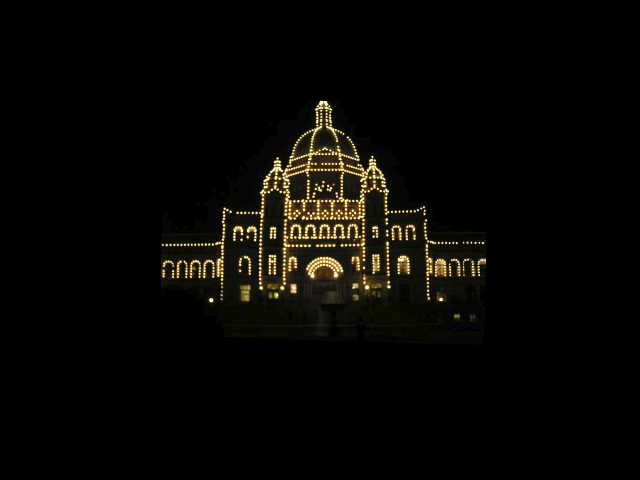}        
    \caption{\small Warped Image}
    \end{subfigure}
    \begin{subfigure}[b]{0.5\linewidth}
    \includegraphics[width=\linewidth]{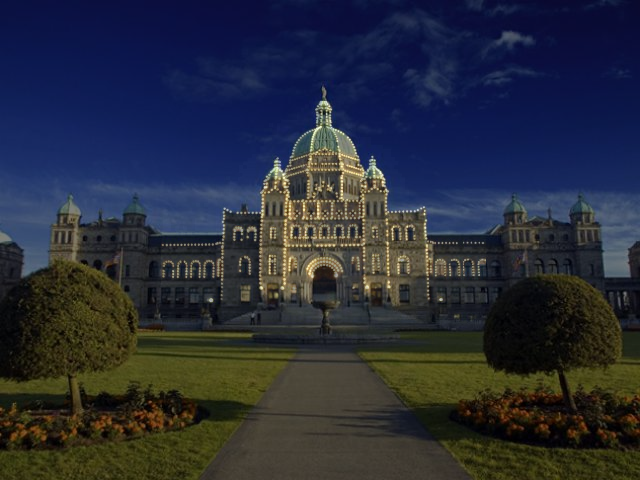}        
    \caption{\small Blended Image}
    \end{subfigure}
    % }
    \caption{Image warping with flows predicted by LIFE.
    }
    \label{Fig: teasor}
\end{figure}
}]
{
  \renewcommand{\thefootnote}%
    {\fnsymbol{footnote}}
  \footnotetext[1]{Zhaoyang Huang and Xiaokun Pan assert equal contributions.}
}

%%%%%%%%% ABSTRACT
\input{00-abstract.tex}

%%%%%%%%% BODY TEXT
\section{Introduction}
\input{01-introduction}
\section{Related Works}
\input{02-relatedworks}

\section{Method}
\input{03-method}

\section{Experiments}
\input{04-experiments}

\section{Conclusion}
\input{05-conclusion}

{\small
\bibliographystyle{ieee_fullname}
\bibliography{egbib}
}

\end{document}

%% file: 00-abstract.tex
\begin{abstract}
We tackle the problem of estimating flow between two images with large lighting variations.
Recent learning-based flow estimation frameworks have shown remarkable performance on image pairs with small displacement and constant illuminations, but cannot work well on cases with large viewpoint change and lighting variations because of the lack of pixel-wise flow annotations for such cases.
We observe that via the Structure-from-Motion~(SfM) techniques, one can easily estimate relative camera poses between image pairs with large viewpoint change and lighting variations.
We propose a novel weakly supervised framework LIFE to train a neural network for estimating accurate lighting-invariant flows between image pairs.
Sparse correspondences are conventionally established via feature matching with descriptors encoding local image contents.
However, local image contents are inevitably ambiguous and error-prone during the cross-image feature matching process, which hinders downstream tasks.
We propose to guide feature matching with the flows predicted by LIFE, which addresses the ambiguous matching by utilizing abundant context information in the image pairs. 
We show that LIFE outperforms previous flow learning frameworks by large margins in challenging scenarios, consistently improves feature matching, and benefits downstream tasks.
\end{abstract}

%% file: 01-introduction.tex
Establishing correspondences between image pairs is a fundamental problem in 3D computer vision, which supports many downstream applications,
such as relative pose estimation~\cite{melekhov2017relative, ummenhofer2017demon}, visual localization~\cite{sarlin2019coarse}, homography estimation~\cite{zhang2020content, le2020deep}, etc.
Pixel-wise dense correspondences further enable image editing~\cite{liu2010sift, luo2020niid}, 3D reconstruction~\cite{agarwal2011building}, etc.
Fully supervised flow estimation methods~\cite{sun2019models,  hui2018liteflownet, teed2020raft} have achieved satisfactory performance for consecutive frames, which usually have similar lighting conditions and pixel displacements between image pairs are usually limited.
In more general scenes, there may exist large viewpoint and illumination changes between pairs of images.
However, annotating dense correspondences by humans for flow estimation is infeasible for such scenarios as the number of pixels is quite large.
To alleviate the impact of large viewpoint change for flow prediction,
there were previous research~\cite{rocco2017convolutional, truong2020glu} that synthesizes the second image by applying geometric transformations to an image.
% \xk{To alleviate the impact of large viewpoint change for flow prediction, previous research~\cite{rocco2017convolutional, truong2020glu} synthesizes the second image by applying geometric transformations to an image.}
Although the accurate dense correspondences with large pixel displacements can be accurately obtained from the geometric transformations in this way,
%and used as training supervisions.
% Supervising the flow estimation neural network .
% we can maintain accurate dense correspondences between two images with large pixel displacements.
%this strategy still cannot handle illumination variations and 
the synthesized images still cannot mimic realistic illumination variations and do not conform to actual geometric transformations.

% but the dense annotations are so expensive especially in more general scenes.
% To address this problem,
% some unsupervised learning and self-supervised learning frameworks~\cite{} are proposed but they rely on either lighting constancy assumption~\cite{}
%  or geometric image transformation~\cite{}, which can not handle large lighting variation.

With the mature Structure-from-Motion~(SfM) technology~\cite{schoenberger2016sfm}, we can collect images of the same scene from the Internet and accurately estimate their camera poses in a unified coordinate system. 
Due to the large time span of these images, the lighting and viewpoint of the images can significantly vary, and their actual variations cannot be easily synthesized via simple image transformations.
%in line with the real world.
%In this way, we can obtain relative poses of the image pairs containing various environmental changes. 
Based on such image pairs with large illumination variations and ground-truth relative pose transformations, we propose the Lighting Invariant Flow Estimation (LIFE) framework that trains a deep neural network to estimate flow with such image-level weak supervisions.
%For achieving flow estimation in more challenge scenarios, we propose LIFE as a novel framework that train the flow estimation neural network with relative camera poses.

% and train the neural network with them through the symmetric epipolar distance loss.

According to epipolar geometry~\cite{yamaguchi2013robust},
the fundamental matrix deduced from the relative camera pose constrain each pixel in the source image to correspond to a line, referred to as epipolar line, in the target image.
If we select a pixel in the target image as the correspondence, the distance from the pixel to the epipolar line is named {\it epipolar distance} and it is expected to be zero, which has been widely used to constrain the searching space of optimization-based optical flow estimation~\cite{valgaerts2008variational, wedel2009structure}.
Besides, once the corresponding pixel is selected, it also corresponds to an epipolar line in the source image, from which we can compute another epipolar distance.
The sum of these two distances is {\it the symmetric epipolar distance}.
DFE~\cite{ranftl2018deep} introduced it as a loss to learn to estimate the fundamental matrix with given correspondences.
Inspired by them, we alternate the learning objective in DFE: via minimizing the symmetric epipolar distance loss, we can train a neural network to estimate dense pixel flows, given an image pair with a given fundamental matrix.
% supervise the flow estimation neural network with the symmetric epipolar distance loss.
Thanks to the adequate training data with different lighting conditions,
% we can learn more robust flow estimation against lighting variation.
our LIFE can learn robust feature representations against 
%estimate flow between two images even with 
severe lighting variations.
% Epipolar distance loss only provides weak signals for flow learning, which allows the predicted flow slide along the corresponding epipolar line, so we generate a synthetic image by applying a transformation to the second image and maintain the location of pixels before and after the transformation as ground-truth flow for accurate flow training.

The proposed symmetric epipolar distance loss only imposes weak constraints on the flow estimation, which allows the predicted flow to slide along the corresponding epipolar line.
To improve the accuracy of flow, we synthesize an image from one image in the image pair with geometric transformations.
The additional pixel-to-pixel constraint derived from the geometric transformations further complements the symmetric epipolar distance loss and achieves accurate lighting-invariant flow estimation.
We present an example that warps an image captured at nighttime to an image captured at daytime via the flows predicted by LIFE~(Fig.~\ref{Fig: teasor}).

% and maintain the location of pixels before and after the transformation as ground-truth flow for accurate flow training.
% To improve the flow accuracy,

% In contrast with sparse features only finding strictly cyclic consistent correspondences, 
% flow learning should take care of each pixel.
% However,
% a pixel in the target image may correspond to multiple pixels in the source image because the foreground object in the target image may occlude some background that is visible in the source image, which denotes the flow is unidirectional and there may be multiple pixels in the source image flow to the same location at the target image.
% Therefore, we only impose cycle consistency on flow pairs that own low cycle distance during training and take the cyclic distance as the correspondence confidence in the test phase.
% Moreover, the multiple pixels in the source image should lie on the epipolar line induced from the target location and inversed relative pose, so we use the symmetric epipolar distance loss.
% According to the formulated image triplet, we unify the symmetric epipolar distance loss and the synthetic transformation loss for achieving robustness against severe appearance change and remaining high flow accuracy.

Besides dense correspondences, establishing sparse correspondences is also an important vision task, which is usually achieved by extracting salient feature points and matching the points according to descriptors.
However, as descriptors of feature points are individually extracted from an image with a limited perception field, descriptors from repeated patterns may be indistinguishable.
Feature points with ambiguous descriptors are easily erroneously matched during inference, which raises the outlier ratio and might impact downstream tasks.
Guiding feature matching with flows is a common strategy in recent visual odometry~\cite{qin2018online}, which alleviates the descriptor ambiguity and reduces computational complexity.
Nonetheless, it is not utilized in other applications such as visual localization as 
previous methods have difficulties in flow estimation for images with large lighting variations.
%achieve remarkable performance in flow estimation between consecutive frames but can not estimate reasonable flow between challenge image pairs.
% In contrast, our LIFE predicts dense correspondences by effectively utilizing context information, which can overcome the local ambiguity.
We propose to improve the sparse feature matching with the flows predicted our proposed LIFE.
To our best knowledge,
we are the first to explicitly address the lighting variation problem of direct flow prediction with the epipolar constraints as weak supervisions and also make use of lighting-invariant flows to improve feature matching in practical scenarios.
With the assist of our LIFE,
the inlier ratio of sparse correspondences increases and the accuracy of the follow-up geometric transformation estimation is significantly improved.

In summary, our proposed approach has the following major contributions: (1) We propose a weakly supervised framework LIFE that trains the flow estimation neural network with camera pose transformations and can predict accurate flow in large viewpoint and lighting variation scenarios. (2) We propose to improve the sparse feature matching with our predicted lighting invariant flows, which is able to alleviate the ambiguity of local descriptor matching. (3) Our proposed LIFE outperforms state-of-the-art flow estimation methods in challenging scenes by large margins and improves sparse feature matching performance to assist downstream applications.

%% file: 02-relatedworks.tex
% \begin{figure*}[t!]
%     \centering
%     \includegraphics[width=1.0\linewidth, trim={10mm 50mm 30mm 30mm}, clip]{LIFE/figures/BiRAFT_Arc.pdf}
%     \caption{Neural network architecture of LIFE. LIFE predicts bidirectional flow $f_{AB}$ and $f_{BA}$ from given image pairs.
%     }
%     \label{Fig:framework}
% \end{figure*}

\noindent \textbf{Learning to estimate flow}
Great efforts have been devoted to estimate flow between image pairs in the past four decades.
With the success of deep neural networks, learning-based flow estimation has achieved impressive results. 
A common optical flow estimation scene is to find pixel correspondences in two consecutive images of a video.
Dosovitskiy~\etal~\cite{dosovitskiy2015flownet} constructed FlowNet, the first end-to-end trainable CNN for predicting optical flow and trained it on the synthetic FlyingChairs dataset.
Since then, a great number of works~\cite{ilg2017flownet, ranjan2017optical, sun2018pwc, sun2019models, hui2018liteflownet, hui2019lightweight, zhong2019unsupervised, teed2020raft} are proposed for improving the neural network architecture.
RAFT~\cite{teed2020raft} achieves the best performance with full supervision in all public optical flow benchmarks by utilizing a multi-scale 4D correlation volume for all pairs of pixels and a recurrent unit iteratively updating the flow estimation. 
Learning flow estimation requires ground-truth pixel-wise annotations, but it is too expensive in real scenes.
Some unsupervised learning~\cite{jason2016back, ren2017unsupervised, meister2018unflow, yin2018geonet, zou2018df, liu2020flow2stereo} frameworks circumvent this requirement by utilizing image warping loss and geometry consistency~\cite{yin2018geonet, zou2018df, liu2020flow2stereo}.
However,
consecutive images under good lighting conditions is a rather simple case and image warping loss assumes constant illumination.
Researchers have recently started to address flow estimation in more challenging scenarios where annotating pixel-wise flow for training becomes even more difficult, such as low-light~\cite{zheng2020optical} or foggy~\cite{yan2020optical} images.
They address the data problem by image synthesize~\cite{zheng2020optical}  or domain transformation~\cite{yan2020optical}.
% Zheng~\etal~\cite{zheng2020optical} synthesized low-light optical flow datasets and directly learn optical flow from low-light noisy images, which enables optical flow estimation in the dark.
% Yan~\etal~\cite{yan2020optical} makes flow estimation practical in foggy scenes with domain transformation techniques.
Large viewpoint and illumination transformation will also significantly increase the difficulty of flow estimation.
\cite{truong2020glu, truong2020gocor, melekhov2019dgc, rocco2017convolutional} focus on improving the neural network architecture for large viewpoint change.
The training data is maintained by applying geometric transformations to images, which is not in line with real scenes.
To alleviate the impact from illumination variation, Simon~\etal~\cite{meister2018unflow} use the ternary census transform~\cite{stein2004efficient} to compensate for additive and multiplicative illumination change but it is a simple handcrafted approximation.
RANSAC-Flow~\cite{shen2020ransac} estimates flow with an off-the-shelf feature extranctor and the RANSAC~\cite{fischler1981random, raguram2012usac} based on multiple homographies, which presents good performance but the resultant flows are retricted by the multiple homographies.
% Meanwhile, they and Yang~\etal~\cite{wang2018occlusion} detect occlusions by predicting bidirectional flow and check the forward-backward consistency.
% We use the RAFT as the neural network architecture and transform it into Bi-RAFT that predicts bidirectional flow between a pair of image.
% We propose to learn flow prediction by the symmetric epipolar distance loss.

\noindent \textbf{Learning with Epipolar Constraints}
% Correspondences between two view are not arbitrarily possible.
Correspondences anchored upon the surfaces of rigid objects between two views must obey the geometric transformation according to the relative pose, which is called epipolar constraints.
As most contents in images are static and rigid, epipolar constraints are widely employed in traditional optimization-based optical flow~\cite{valgaerts2008variational, wedel2009structure, yamaguchi2013robust}.
Recently, epipolar constraints are introduced to construct loss functions for learning.
Subspace constraint and low-rank constraint~\cite{zhong2019unsupervised} are applied in unsupervised optical flow learning.
% are derived from epipolar constraints.
Some local feature learning methods leveraged epipolar constraints to train detectors~\cite{yang2019learning} and descriptors~\cite{wang2020learning}.
Ranftl~\etal~\cite{ranftl2018deep} learned to estimate fundamental matrix from given correspondences with the symmetric epipolar distance loss.
Reducing the correspondence searching space to a narrow band along the epipolar line~\cite{he2020epipolar} is also a common strategy.
To our best knowledge, we are the first that directly learn flow by the symmetric epipolar distance loss.
Fueled by abundant and various training data maintained from SfM, LIFE is robust to large viewpoint and illumination change.

%% file: 03-method.tex
We propose the weakly supervised lighting invariant flow estimation (LIFE) framework, which only requires whole-image camera pose transformations as weak supervisions and shows effectiveness on establishing accurate and dense correspondences between images with significant lighting and viewpoint variations.
%and a LIFE-based sparse correspondence establishment algorithm.
In this section, we  will elaborate the LIFE framework and the LIFE-based sparse correspondences establishment.
The neural network architecture of RAFT~\cite{teed2020raft} has been proven successful in learning optical flows but it only learns to predict flows from a source image to a target image. 
We use RAFT to predict flows in both source-to-target and target-to-source directions with shared parameters as our training losses are applied in both directions.
Training data are prepared as image pairs with corresponding fundamental matrices, which can be easily obtained from the Ineternet and processed by SfM techniques to recover the camera poses.
%We then discuss training the flow network with these image pairs and improve it by formulating 
Because of the weak supervisions from whole-image camera poses, we further regularizing the flow learning with cycle consistency and synthetic dense flows.
Finally, we show how to establish highly accurate sparse correspondences with the bidirectional flows predicted by LIFE.

\begin{figure}[t!]
    \centering
    \includegraphics[width=0.9\linewidth, trim={5mm 35mm 0mm 10mm}, clip]{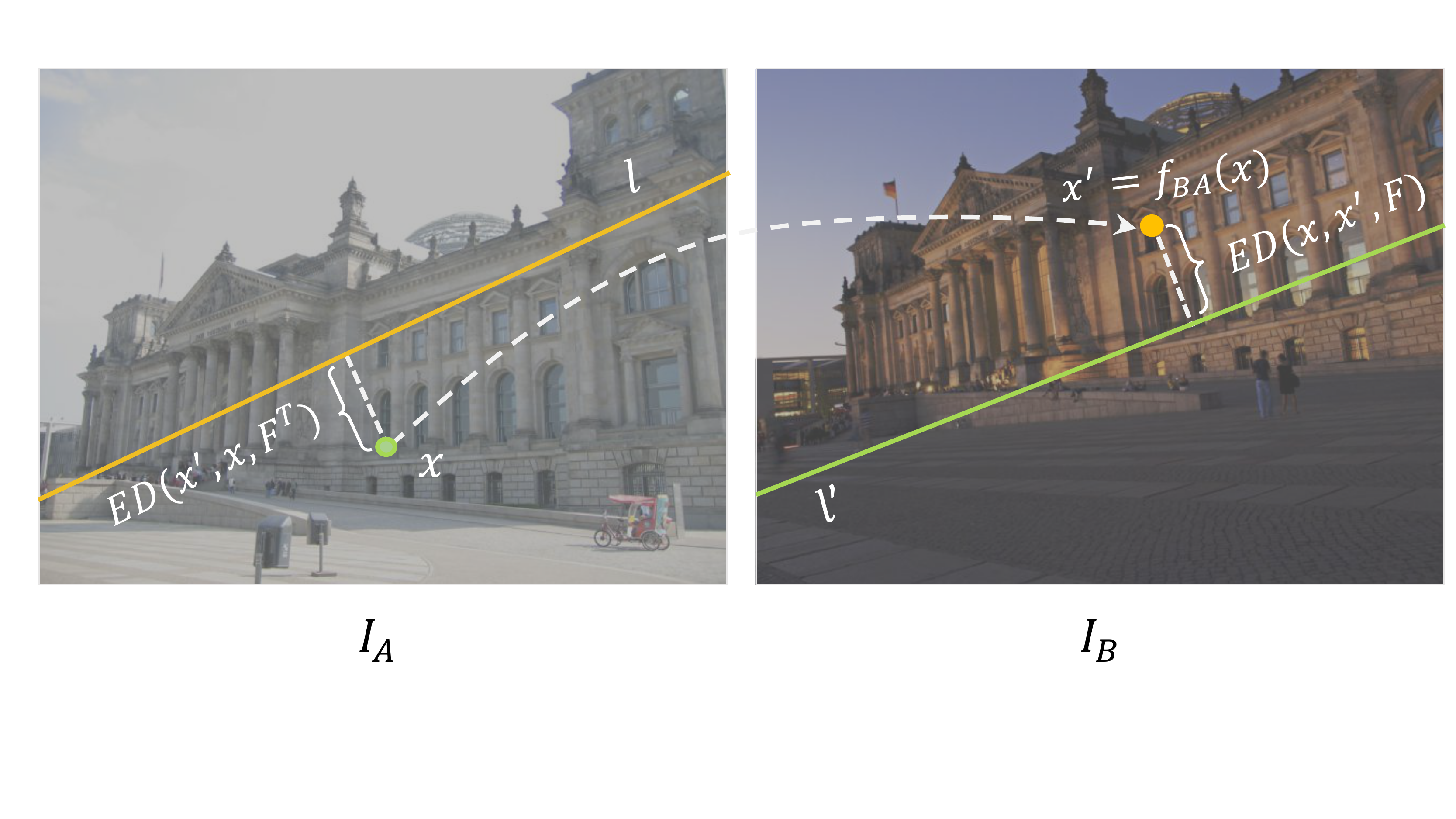}
    \caption{Symmetric epipolar distance. 
    }
    \label{Fig:SED}
    \vspace{-3mm}
\end{figure}

% The critical step is to collect data satisfying these conditions with corresponding labels for training.
% Based on SfM technique, we can maintain a large amount of satisfactory image pairs with their relative camera poses.
% We first introduce the symmetric epipolar distance loss, which weakly supervise the flow estimation neural network by relative camera poses.
% Weak constraints can not generate accurate supervisory signals, so we formulate an image triplet given an image pair, which complements the 

\subsection{Training Light-invariant Dense Flows from Camera Poses}

As camera poses between images with large appearance and geometric variations are easier to acquire than dense flows, our proposed LIFE learns from such camera poses as weak supervisions to predict lighting invariant dense flows.
% With the bidirectional flows output by the network, we propose a symmetric epipolar distance (SED) loss to achieve this goal.

\vspace{3pt}
\noindent \textbf{Symmetric Epipolar Distance (SED) loss.} Based on the epipolar geometry, we propose a symmetric epipolar distance (SED) loss to achieve this goal.
We compute the fundamental matrix $\mathbf{F}$ from $I_A$ to $I_B$ according to their camera intrinsic parameters and relative camera pose.
As shown in Fig.~\ref{Fig:SED},
$\mathbf{F}$ restricts that a pixel location $\mathbf{x}\in \mathbb{R}^2$ in image $I_A$ can only be mapped to one of the points on a line $l'=\mathbf{Fx}$, referred to as {\it epipolar line} in image $I_B$.
For each point $\mathbf{x}$ in image $I_A$, our network estimates its corresponding point in image $I_B$ as $\mathbf{x}'=f_{B\leftarrow A}(\mathbf{x})$.
If the $A$-to-$B$ flows are ideal, the distance of the corresponding pixel location $x'$ to the epipolar line $l'$, named epipolar distance~(ED), shall be zero.
% With the static scene assumption, the corresponding pixel $x'$ should lie on $l'$, which holds $x'^T\mathbf{F}_{BA}x=0$.
% Bi-RAFT should minimize the pixel distance of $x'$ to $l'$,. 
% Symmetrically, $x$ is also supposed to lie on the epipolar line $l=\mathbf{F}^Tx'$ derived from $x'$. 
Reversely, $x$ is also supposed to lie on the epipolar line $l$ derived from $x'$ in image $I_B$ if the reverse flows are ideal, and the distance from $x$ to the epipolar line $l$ should also be zero.
The sum of the two epipolar distances is defined as the symmetric epipolar distance~(SED).
According to the epipolar geometry, the inverted fundamental matrix equals the transpose of the fundamental matrix, we can therefore compute the SED as
\begin{equation}
SED(\mathbf{x}, \mathbf{x}', \mathbf{F})
= ED(\mathbf{x}, \mathbf{x}', \mathbf{F}) + ED(\mathbf{x}', \mathbf{x}, \mathbf{F}^T).
\end{equation}
% $$
% ED = |x'^T\mathbf{F}_{BA}x\left(\frac{1}{l'_{1}^{2}+l_{2}'^{2}}\right)|
% $$
% Given a pair of images $I_A$ and $I_B$ with their camera intrinsic parameters $K_A$ and $K_B$ and the relative camera pose $\mathbf{R}_{BA}, \mathbf{t}_{BA}$ from $I_A$ to $I_B$, we can compute the fundamental matrix by:
% $$
% \mathbf{F}_{BA} = \mathbf{K}_B^{-T}[\mathbf{t}_{BA}]_\times \mathbf{R}_{BA}\mathbf{K}_A^{-1}
% $$

Given the $A$-to-$B$ dense flows $f_{B\leftarrow A}$ predicted by our flow network, we can define the following SED loss to evaluate their accuracies by computing SED for all flows in $f_{B\leftarrow A}$.
% , which gradually increases if the $f_{B\leftarrow A}$ flows deviate from their ground-truth.
%
% According to epipolar geometry~\cite{}, the fundamental matrix $\mathbf{F}_{BA}$ maps a pixel $x$ in the image $I_A$ to a line $l'=\mathbf{F}_{BA}x$ in the image $I_B$.
% With static scene assumption, the corresponding pixel $x'$ should lie on $l'$, which holds $x'^T\mathbf{F}_{BA}x=0$.

% \begin{equation}
% L_{sym}(\mathbf{x}_i, \mathbf{x}'_i, \mathbf{F})
% = \mathbf{x}_i^T\mathbf{F}\mathbf{x}_i'\left(\frac{1}{\left(\mathbf{Fx}_{i}\right)_{1}^{2}+\left(\mathbf{Fx}_{i}\right)_{2}^{2}}+\frac{1}{\left(\mathbf{F}^{\top} \mathbf{x}_{i}^{\prime}\right)_{1}^{2}+\left(\mathbf{F}^{\top} \mathbf{x}_{i}^{\prime}\right)_{2}^{2}}\right)
% \end{equation}
%
\begin{equation}
L_{SED}
= \sum_{\mathbf{x}_i \in S}SED(\mathbf{x}_i, f_{B\leftarrow A}(\mathbf{x}_i),\mathbf{F}),
\end{equation}
where $S$ is the set containing all pixel locations in $I_A$.
Compared with photometric consistency loss used in existing unsupervised flow learning frameworks, which assumes constant lighting conditions, the proposed epipolar distance loss is only determined by the fundamental matrix (or relative camera pose). Therefore, the SED loss works even when there exist significant lighting variations between the two images.

\noindent \textbf{Cycle consistency regularization.} The cycle loss measuring flows' cycle consistency $d(\mathbf{x}) = ||f_{A\leftarrow B}(f_{B\leftarrow A}(\mathbf{x})) - \mathbf{x}||_2$ is a common regularization term in correspondence learning~\cite{rocco2018neighbourhood}.
However, pixels in occluded regions do not satisfy the cycle consistency assumption and might infer large cycle distance errors to overwhelm the cycle loss. Therefore, making all flows to be cycle consistent would actually hinder the training and generate over-smooth flow fields with degraded performances.
%a pair of unreliable flow cycle consistent is useless, which will generate large cycle distance error and overwhelm the cycle loss. Therefore, directly adopting the cycle loss for all pixels will generate over-smooth flow and even achieve degraded performance.
% but minimizing the cycle distance of bidirectional flow for occluded areas is improper and make useless efforts with unreliable prediction, which will result in over-smoothness.
Inspired by unsupervised optical flow learning~\cite{meister2018unflow, yin2018geonet}, we filter out pixels with too large cycle distance errors and use the cycle loss from the kept pixels.
$$
L_{cyc} = \sum_{\mathbf{x}_i \in S}\mathbf{1}(d(\mathbf{x}_i) \leq \max\{\alpha, \beta ||f_{B\leftarrow A}(\mathbf{x}_i)||_2\})d(\mathbf{x}_i),
$$
where $\mathbf{1}$ denotes the indicator function. A pixel $\mathbf{x}$ whose cycle distance is larger than $\alpha$ and $\beta ||f_{B\leftarrow A}(\mathbf{x})||_2$ will be filtered in the cycle loss.

\begin{figure}[t!]
    \centering
    \includegraphics[width=0.9\linewidth, trim={10mm 10mm 50mm 0mm}, clip]{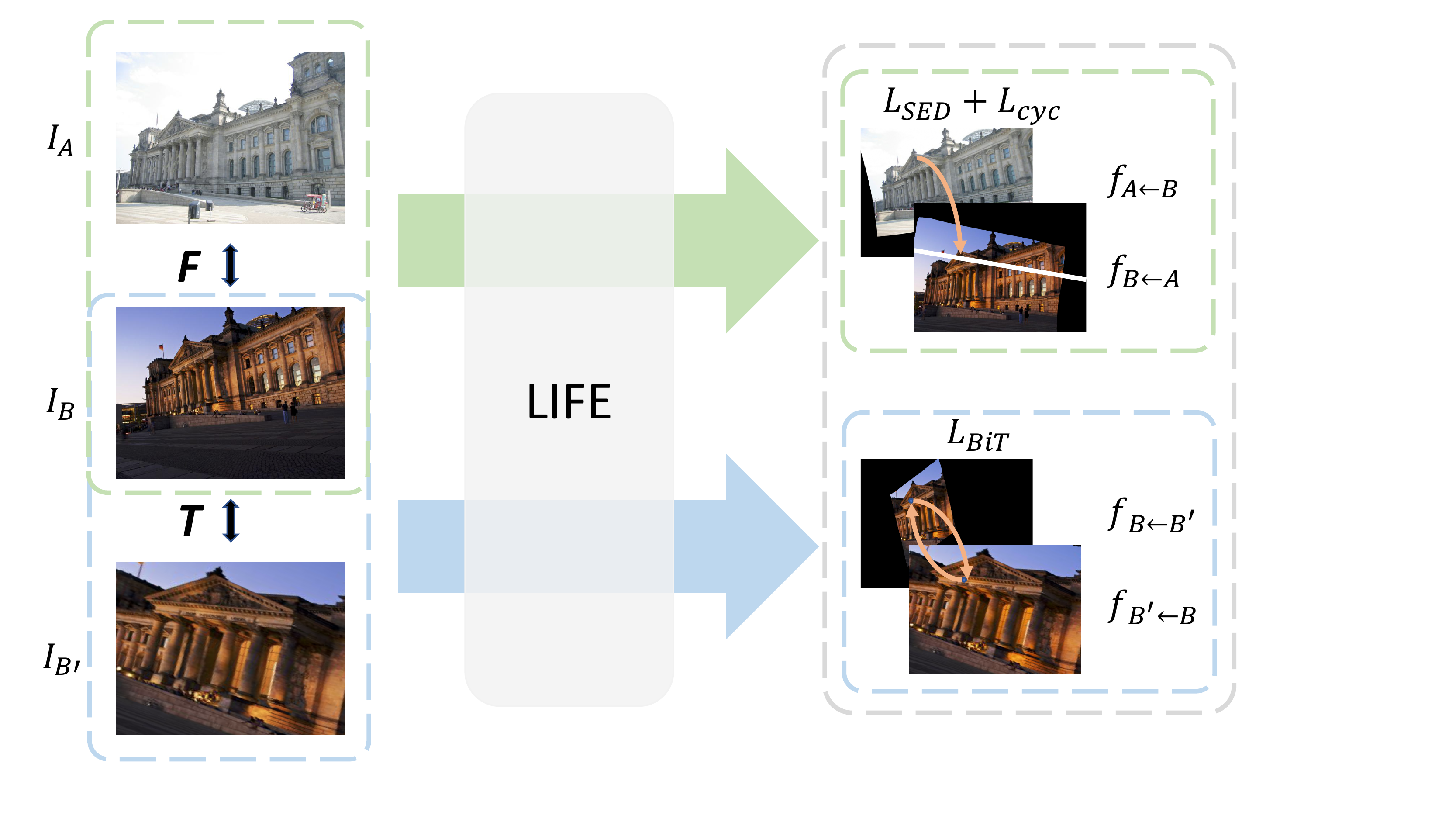}
    \caption{The overall training loss. The top pair with large lighting and viewpoint variation is supervised by the proposed Symmetric Epipolar Distance (SED) loss and the cycle consistency loss. The bottom pair with a synthetic transformation is supervised by the bidirectional geometric transformation (BiT) loss.
    }
    \label{Fig:Triplet}
\end{figure}

% \subsection{Image Triplet Formulation}
\vspace{3pt}
\noindent {\bf Synthetic dense-flow regularization.} Although the SED loss can work on image pairs of actual scenes with significant lighting and pose variations, it can only provide weak supervision on minimizing the distances from points to their epipolar lines. In other words, as long as the predicted flow aligns points to their epipolar lines in the other image, their SED loss is minimal. However, points' ground-truth correspondences should also be single points. 
%can predict flow that can roughly align images but lack accuracy.
In order to improve the accuracy of the flow prediction, we propose to regularize the estimated flows with synthetic pixel-to-pixel supervisions.
Inspired by Rocco~\etal~\cite{rocco2017convolutional},
for each image pair, we randomly generate an affine or thin-plate spline transformation $\mathbf{T}$ to transform image $I_B$ of the image pair.
% On the contrary, Rocco~\etal~\cite{} synthesize images from real images with generated affine and thin-plate spline transformations, which can produce pixel-to-pixel flow constraints.
% Inspired by them, we synthesize the third image $I_{B'}$ from the second image $I_B$ in an image pair with a geometric transformation $\mathbf{T}$.
In this way, we can create a synthesized image $I_{B'}$ and a synthesized image pair $<I_B, I_{B}'>$~(Fig.~\ref{Fig:Triplet}) with accurate dense pixel-to-pixel correspondence ground truth.
Given the location of a pixel in $I_{B}$, we can compute its accurate corresponding location in $I_{B'}$ according to the synthesized geometric transformation $\mathbf{T}$, and vice versa via the inverse of the geometric transformation $\mathbf{T}^{-1}$.
% We therefore additionally supervise our flow network with the accurate pixel-to-pixel dense correspondences but without any lighting variation.
We therefore additionally regularize our flow network with a bidirectional geometric transformation~(BiT) loss.
The BiT loss supervises the bidirectional flow with accurate pixel-to-pixel dense correspondences, but the image pair has a synthetic image transformation and zero lighting variation,
\begin{equation}
\begin{split}
L_{BiT} = & \sum_{\mathbf{x}_i\in S_{B}}||f_{B'\leftarrow B}(\mathbf{x}_i) - \mathbf{T}(\mathbf{x}_i)||_1 + \\
&\sum_{\mathbf{x}_i\in S_{B'}}||f_{B\leftarrow B'}(\mathbf{x}_i) - \mathbf{T}^{-1}(\mathbf{x}_i)||_1,
\end{split}
\end{equation}
where $S_{B}$ and $S_{B'}$ contains locations of valid pixels in $I_{B}$ and $I_{B'}$ respectively. For points that might be out of the visible region in the target images, we regard these flow as invalid and filter out them for loss computation.

% \vspace{3pt}
\noindent \textbf{The overall loss.} For image $I_B$ and its synthetically transformed version $I_{B'}$,
$I_B$ is related to $I_A$ according to their fundamental matrix $\mathbf{F}$, while the dense correspondences between $I_B$ and $I_{B'}$ are precisely determined by the synthesized geometric transformation $\mathbf{T}$.
We apply the SED loss and the cycle loss to the bidirectional flow deduced from $I_A$ and $I_B$, and the BiT loss to the bidirectional flow deduced from $I_B$ and $I_{B'}$~(Fig.~\ref{Fig:Triplet}).
The SED loss supervises flows between actual image pairs with natural lighting and viewpoint variations but only provides weak supervision signals.
On the contrary, the BiT loss regularizes flow with strict dense correspondences with constant lighting conditions and synthesized image transformation.
%the lighting is constant and one image is synthesized.
Unifying the losses of the images can simultaneously mitigate both their drawbacks and contribute to training a robust lighting-invariant flow estimation network.
%: predicting accurate flows between image pairs that occur significant lighting and viewpoint variations.

\subsection{Finding Sparse Correspondences with LIFE}

Dense correspondences can be used in many applications but are not mandatory in whole-image geometric transformation estimation tasks, such as relative pose estimation and homography estimation.
In such tasks, we only need a small number of sparse but accurate correspondences to estimate whole-image transformation.
Sparse correspondences can be generally obtained by detecting locally salient feature points in the image pair with feature descriptors and establishing correspondences between feature points with similar descriptors.
However, descriptors encode contents of local image patches, which are inevitably ambiguous in terms of the global image context.
Erroneous matches caused by ambiguous descriptors is a long-standing problem even with learning-based descriptors, which both reduce inlier ratio and the number of effective correspondences for the follow-up transformation estimation step.
In contrast, our dense flows are less possible to be trapped by local ambiguous patterns thanks to the context information.
We design a simple but effective two-stage algorithm to identify sparse correspondences.
% We believe the flows predicted by LIFE are confident, so we only match features in local regions guided by the predicted flows in the stage 1 .
In stage 1, we identify correspondences with the assist of LIFE.
We believe the flows predicted by LIFE are confident, so we only need to find the corresponding features in local regions guided by the predicted flows.
The correspondences identified in stage 1 are of high quality but the correspondence number may be unsatisfactorily low because of the strict flow constraints.
Therefore, in stage 2, we try to identify more correspondences via the remaining feature points of stage 1.
% can still potentially supplement more useful correspondences from them simply with descriptors.
% Therefore, we try to identify more correspondences for the failed features based only on descriptors in the stage 2.
% The quality of correspondences identified in the stage 2 have lower quality but it ensures satisfactory match number the cases
%in terms of whether using sparse local features.
% Based on LIFE, we propose three simple but effective algorithms that gradually integrated local features into LIFE for sparse feature establishment.

\begin{figure}[t!]
    \centering
    \includegraphics[width=0.9\linewidth, trim={0mm 80mm 60mm 5mm}, clip]{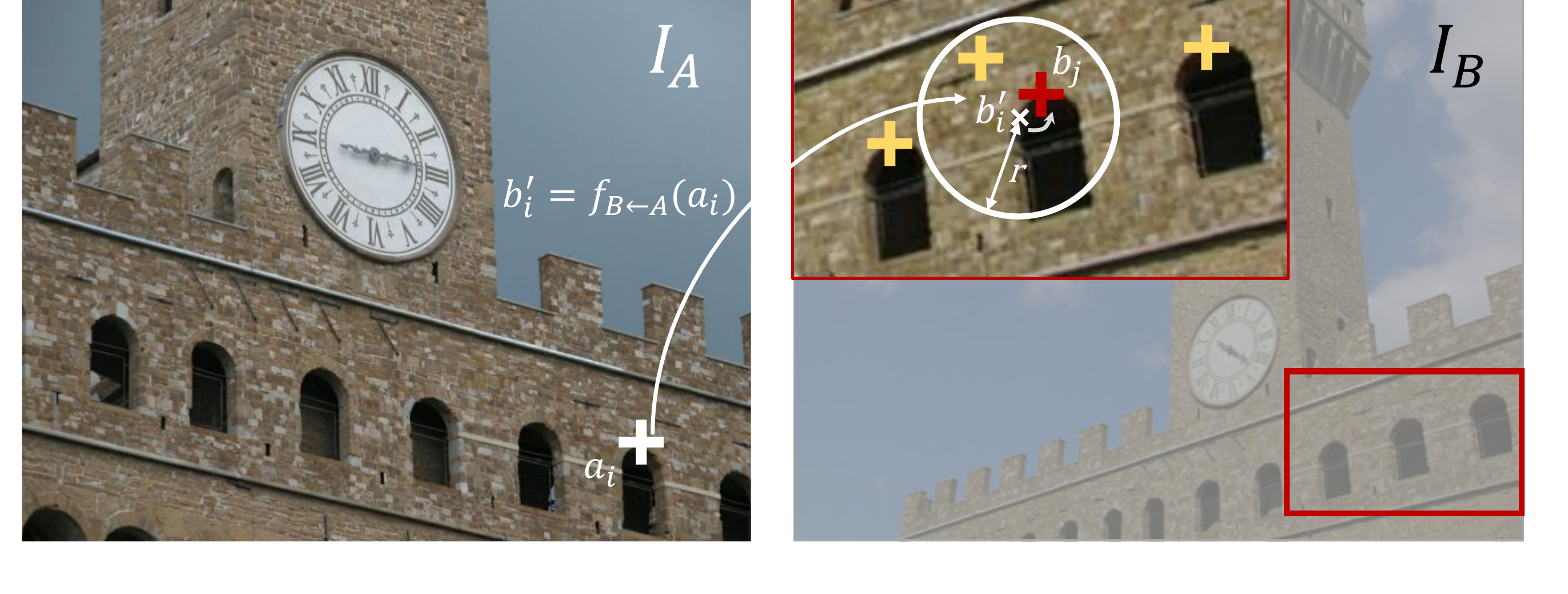}
    \caption{With the flow predicted by LIFE, we find the feature point $\mathbf{b}_j$ whose descriptor is the closest to that of $\mathbf{a}_i$ in the circle as the corresponding point. 
    }
    \label{Fig: GM}
    \vspace{-3mm}
\end{figure}

\noindent \textbf{Stage 1.}
We denote the sets of sparse feature points detected in $I_A$ and $I_B$ by $\mathbf{a}_i \in A$ and $\mathbf{b}_j \in B$, and denote their descriptors by $\mathbf{q}^a_i$ and $\mathbf{q}^b_j$.
For a given query feature point $\mathbf{a}_i$, we first calculate its warped point $\mathbf{b}'_i = f_{B\leftarrow A}(\mathbf{a}_i)$ in $I_B$ according to the predicted flows $f_{B\leftarrow A}$.
Inside the circle centered at $\mathbf{b}_i'$,
we regard the feature point $\mathbf{b}_j$ whose descriptor is the closest to that of $\mathbf{a}_i$ as the corresponding point~(Fig.~\ref{Fig: GM}), \ie,
% We regard a feature point $\mathbf{b}_j$ in image $I_B$ as its corresponding point if 1) $\mathbf{b}_j$ locates in a circle centered at $\mathbf{b}'_i$, and 2) $\mathbf{b}_j$'s descriptor is the closest to the that of $\mathbf{b}_i$'s, \ie,
\begin{equation}
\centering
\begin{aligned}
% &\mathbf{y}_j = \arg \min  q^x_i^Tq^y_j \\
% &\text{subject to} ||\mathbf{y}_j - \mathbf{x}'_i||_2 \leq r,
&j = \underset{j}{\arg \max} \,\, \mathbf{q}^{a}{}_{i}^T\mathbf{q}^{b}_{j}, \\
&\mathrm{s.t.}~ ||\mathbf{b}_j - \mathbf{b}'_i||_2 \leq r.
\end{aligned}
\end{equation}
$r$ is the radius of the circle, which is set as 5 pixel.
In this way, we can try to identify a corresponding feature point $\mathbf{b}_j$ in $I_B$ for each $\mathbf{a}_i$ in $I_A$, and create a matching feature point set $M_{B\leftarrow A}$ that satisfy the above formula. Reversely, we can use the same strategy to establish a matching point set $M_{A\leftarrow B}$ in the reverse direction from $I_B$ to $I_A$.
%By narrowing the searching radius of matching with the guidance of flow, some matches that would be erroneously matched with only descriptors caused by the ambiguity of descriptors can be corrected.
%We denote these two set of matches by $M_{y\leftarrow x}$ and $M_{x\leftarrow y}$.
%As the matching process is unidirectional,
%the matched feature of $\mathbf{y}_j$ who is the matched feature of $\mathbf{x}_i$ may not be $\mathbf{x}_i$, \ie, $\mathbf{x}_i \neq M_{x\leftarrow y}(M_{y\leftarrow x}(\mathbf{x}_i))$, which are undistinguished and would be pruned.
%The survived features with their matches are ensured with $\mathbf{x}_i = M_{x\leftarrow y}(M_{y\leftarrow x}(\mathbf{x}_i))$, which is called as the mutual match check. The correspondences derived from them are the outcome in the first stage, and the other features are remained for the next stage.
Given the two matching point sets of the two matching directions, we only keep the final matching point pairs as those survive the cycle consistency check, \ie, $\mathbf{x}_i = M_{A\leftarrow B}(M_{B\leftarrow A}(\mathbf{x}_i))$

\noindent \textbf{Stage 2.} 
We collect the features that are weeded out in stage 1 and denote them by $A_w$ and $B_w$ according to which image they belong to. More correspondences are tried to be established from them as supplements. 
Given a feature point $\mathbf{a}_i \in A_w$ as a query feature, we directly find the feature point $\mathbf{b}_j \in B$ who has the closest descriptor in $I_B$ as its corresponding point.
Note that this not equivalent to establishing correspondences between all points $A$ and $B$ with feature descriptors, as $A_w$ and $B_w$ are smaller feature point sets after the stage-1 matching.
%We believe the correspondences identified by our flow in stage 1 are highly confident so we would directly filter out $\mathbf{a}_i$ if $\mathbf{b}_j \in B - B_w$, \ie,  $\mathbf{b}_j$ has been successfully identified in stage 1.
Similar to stage 1, 
%we try to find the corresponding points for all $\mathbf{a}_i \in A_w$ and $\mathbf{b}_j \in B_w$ and prune 
only the matched point pairs that satisfy the cycle consistency check would eventually be kept.
The survived matches are the outcome correspondences in stage 2.

Directly matching local features with descriptors would cause erroneous matches due to the ambiguity of descriptors, which can be corrected by the local match in stage 1 with the guidance of the flows, so the quality of the correspondences highly relies on the flows.
Unreliable flows can produce misleading guidance, which may even impact the matching.
As LIFE is able to predict accurate flows in challenging scenarios, we can identify more inlier correspondences through this algorithm.

%% file: 04-experiments.tex
\begin{table*}[t]
    \centering
    \resizebox{1.0\linewidth}{!}{
    \begin{tabular}{c|ccccc|ccccc}
    % \hline
     & \multicolumn{5}{c|}{Viewpoint} & \multicolumn{5}{c}{Illuimination~(trans)}\\
     & I & II & III & IV & V & I & II & III & IV & V \\
     \hline 
     CAPS~\cite{wang2020learning} & 34.7/72.4 & 65.2/56.9 & 76.7/53.8 & 84.7/48.5 & 107.6/38.5 & 26.8/66.7 & 55.6/52.8 & 66.0/48.2 & 78.1/43.6 & 104.5/35.7 \\
     RAFT~\cite{teed2020raft} & 15.2/78.0 & 60.5/49.0 & 61.8/47.0 & 67.8/34.8 & 95.8/25.3 & 32.6/58.7 & 93.6/26.9 & 103.7/25.7 & 100.6/22.0 & 141.4/17.3 \\
     DGC~\cite{melekhov2019dgc} & 5.5/78.4 & 13.4/66.1 & 23.3/61.9 & 38.9/51.0 & 21.8/41.6 & 11.5/58.3 & 23.0/41.6 & 30.3/32.3 & 28.9/29.0 & 45.7/23.9 \\
     GLU~\cite{truong2020glu} & $\mathbf{0.9}$/98.6 & 8.1/91.0 & 17.1/84.5 & 19.5/81.0 & 32.6/67.6 & 4.8/92.5 & 15.2/79.7 & 23.3/76.8 & 18.0/74.1 & 37.8/59.4 \\
    \hline
    LIFE &  $\mathbf{0.9}$/$\mathbf{98.8}$ & $\mathbf{6.0}$/$\mathbf{94.6}$ & $\mathbf{5.3}$/$\mathbf{94.2}$ & $\mathbf{5.3}$/$\mathbf{91.6}$ & $\mathbf{11.9}$/$\mathbf{88.0}$ & $\mathbf{1.3}$/$\mathbf{95.5}$ & $\mathbf{5.8}$/$\mathbf{87.9}$ & $\mathbf{13.8}$/$\mathbf{82.9}$ & $\mathbf{10.4}$/$\mathbf{85.9}$ & $\mathbf{14.6}$/$\mathbf{76.5}$\\
    \hline
    \end{tabular}
    }
    \caption{Flow on HPatches. We compare the AEPE/Acc~(\%). The difficulty gradually increases from I to V.
    }
    % \vspace{-3mm}
    \label{tab: hpatches pck}
\end{table*}

\begin{table}[t]
    \centering
    \resizebox{0.9\linewidth}{!}{
    \begin{tabular}{ccccc}
    \hline
     & \multicolumn{2}{c}{KITTI-2012} & \multicolumn{2}{c}{KITTI-2015}\\
     & AEPE & F1 & AEPE & F1 \\
    \hline 
    % LiteFlowNet~\cite{} & 4.00  & 17.47 & 10.39 & 28.50 \\
    % PWC-Net~\cite{} & 4.14  & 20.28 & 10.35 & 33.67 \\
    DGC~\cite{melekhov2019dgc} & 8.50 & 32.28 & 14.97 & 50.98\\
    GLU~\cite{truong2020glu} & 3.34 & 18.93 & 9.79 & 37.52\\
    GLU$^+$~\cite{truong2020gocor} & 3.14 & 19.76 & 7.49 & 33.83\\
    GLU-GOC$^+$~\cite{truong2020gocor} & 2.68 & 15.43 & $\mathbf{6.68}$ & 27.57\\
    % HD$^3$~\cite{} & 6.85 & 35.47 & 16.87 &  49.93\\
    UFlow~\cite{49244} & - & - & 9.40 & - \\
    DDFlow~\cite{liu2019ddflow} & 8.27 & - & 17.26 & - \\
    R-Flow~\cite{shen2020ransac} & - & - & 12.48 & - \\
    \hline
    LIFE & $\mathbf{2.59}$ & $\mathbf{12.94}$ & 8.30 & $\mathbf{26.05}$ \\
    \hline
    \end{tabular}
    }
    \caption{Generalization of optical flow on KITTI. $^+$ denotes that they use the \textit{dynamic} training strategy.
    }
    \vspace{-3mm}
    \label{tab: kitti flow}
\end{table}

The core contributions of LIFE are making flow estimation between images with challenging lighting and viewpoint variations practical and addressing the ambiguity problem of sparse feature matching with flows .
To demonstrate the effectiveness of LIFE, we compare LIFE with state-of-the-art methods in flow estimation, sparse correspondence establishment, and downstream geometric model estimation.
We select R2D2~\cite{NEURIPS2019_3198dfd0} as the representative local feature for the LIFE-based sparse correspondence identification~(LIFE+R2D2).
At the end, we perform an ablation study to investigate individual modules and the improvement of LIFE for other local features.
More interesting experiments and ablation studies are provided in the supplementary materials and are {\it strongly recommended}.

\noindent \textbf{Datasets.} 
The MegaDepth dataset~\cite{li2018megadepth} collects images under various viewpoint and lighting conditions of 196 scenes and computes their poses with the SfM technique.
We solely train LIFE on MegaDepth with camera poses.
% Besides camera poses, the MegaDepth dataset also contains sparse correspondences between image pairs thanks to the SfM technique.
% Therefore, we also evaluate sparse flows and relative pose estimation on the MegaDepth.
The scenes used for evaluation have been excluded from training.
The HPatches dataset~\cite{balntas2017hpatches} collects image sequences of different scenes under real condition variation.
The image sequences can be divided into two subsets: \textit{Viewpoint} captures images with increasing viewpoint change but under the same illumination condition; \textit{Illumination} captures images under increasing illumination but with almost the same viewpoint.
Each image sequence contains a reference image and 5 source images with increasing variation level.
The ground truth homography is provided for all images, so we qualitatively evaluate all the dense correspondences, sparse correspondences, and homography estimation with controlled conditions.

\subsection{Flow Evaluation}

We evaluate our flows on the KITTI 2012 flow~(training), KITTI 2015 flow~(training)~\cite{geiger2013vision}, Hpatches~\cite{balntas2017hpatches}, RobotCar~\cite{maddern20171, larsson2019cross} and MegaDepth datasets~\cite{li2018megadepth}.
KITTI datasets annotate dense flows between consecutive images, which only have small motion and constant illumination.
The KITTI 2015 flow contains more dynamic objects than the KITTI 2012 flow.
% LIFE shows strong competitiveness on KITTI even it does not target on it.
HPatches contains image pairs with viewpoint and illumination variations of different levels.
The dense correspondences used in evaluation is computed from the provided homography.
However, these two datasets have certain limitations.
To compare the flow prediction performance in real scenes, we further evaluate LIFE on RobotCar~\cite{maddern20171, larsson2019cross} and MegaDepth datasets~\cite{li2018megadepth}.
Dense correspondence annotations are not available in such challenging scenes with large lighting and viewpoint variations.
We therefore evaluate the flow performance on provided sparse correspondences.

\noindent \textbf{Dense Flow Estimation on KITTI.}
We first test flow estimation on the lighting-constant KITTI.
We compare LIFE with state-of-the-art unsupervised flow learning methods that are not fine-tuned on KITTI.
Following the standard optical flow evaluation protocol, we use the Average End-Point Error~(AEPE) and F1 scores in Tab.~\ref{tab: kitti flow}.
LIFE achieves the best performance on the KITTI 2012 flow.
Compared with the KITTI 2012 flow, the KITTI 2015 flow contains more dynamic objects.
Truong \etal~\cite{truong2020gocor} introduced a \textit{dynamic} training strategy by augmenting images with random independently moving objects from the COCO~\cite{lin2014microsoft} so they achieve lower AEPE than our LIFE on the KITTI 2015 flow.
Nonetheless, LIFE still has smaller F1 error even without the {dynamic} training strategy.
Moreover, the experiments on KITTI can only demonstrate the flow estimation performance on cases of simple lighting-constant consecutive images, while LIFE focuses on addressing flow estimation in challenging lighting- and viewpoint-varying scenarios, which will be demonstrated in the following experiments.

\begin{figure*}[t]
\hsize=\textwidth
    \centering
    \begin{subfigure}[c]{0.7\linewidth}
        \includegraphics[width=\linewidth, trim={20mm 0mm 20mm 0mm}, clip]{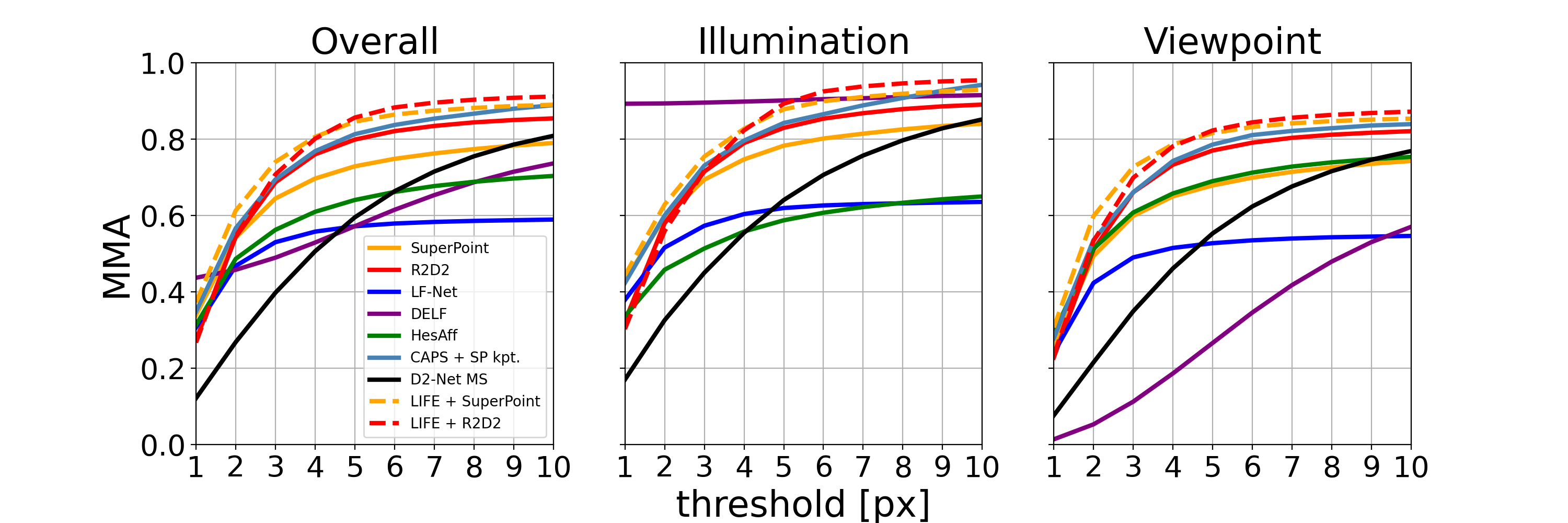}   
    \end{subfigure}
    \begin{subfigure}[c]{0.25\linewidth}
        \centering
    \resizebox{1.0\linewidth}{!}{
    \begin{tabular}{ccc}
    \hline
     & \#Feat & \#Match \\
    \hline 
    D2-Net MS~\cite{Dusmanu2019CVPR} & 8.3K & 2.8K \\
    LF-Net~\cite{ono2018lf} & 0.5K & 0.2K \\
    DELF~\cite{noh2017large} & 4.6K & 1.9K\\
    HesAff~\cite{mikolajczyk2004scale, arandjelovic2012three} & 6.7K & 2.8K \\
    CAPS+SP kpt.~\cite{wang2020learning} & 4.4K & 1.5K  \\
    SuperPoint \cite{detone2018superpoint} & 1.7K & 0.9K\\
    R2D2~\cite{NEURIPS2019_3198dfd0} & 5.0K & 1.8K \\
    LIFE+SP & 1.7K & 1.0K \\
    LIFE+R2D2 & 5.0K & 2.1K \\
    \hline
    \end{tabular}
    }
    \end{subfigure}
    \caption{Sparse correspondence identification on HPatches.
    }
    \label{Fig: mma}
    \vspace{-3mm}
\end{figure*}

\noindent \textbf{Dense Flow Estimation on Hpatches.}
We compute the AEPE and accuracy of compared methods in Tab.~\ref{tab: hpatches pck} with increasing levels of difficulty~(from I to V).
As the image pairs in the \textit{Illumination} subset share the same viewpoint, the ground truth flow is $\mathbf{0}$ for all pixels, which is too simple.
We augment the target images in the \textit{Illumination} subset with generated homographies, so the ground truth flows are no longer all-zero flows.
Here, the accuracy is calculated as the percentage of pixels whose endpoint errors are smaller than 5.
CAPS~\cite{wang2020learning} learns dense descriptors for feature matching, which establishes dense correspondences by finding the nearest descriptors.
It has low accuracy because it does not utilize context information.
RAFT~\cite{teed2020raft} is the state-of-the-art supervisory flow estimation method, which fails as well when encountering difficult cases~(II-V).
DGC-Net~\cite{melekhov2019dgc} and GLU-Net~\cite{truong2020glu} are trained by synthesized images.
LIFE consistently outperforms all previous methods and presents increasing superiority from I to V.
In Viewpoint~(V) and Illumination~(V), which include the most challenging cases, LIFE reduces 63.5\% and 61.4\% AEPE, and raises 20.4\% and 17.1\% accuracy.
These experiments show the remarkable performance of LIFE in the difficult cases with illumination and viewpoint variations.

\noindent \textbf{Sparse Flow Estimation on RobotCar and MegaDepth.}
% We evaluate our method's capability on estimating sparse flows 
These two datasets contain image pairs that correspond to various conditions such as dawn and night.
% but image pairs in the RobotCar dataset share similar viewpoints. 
%To compare the flow estimation performance in challenging real scenes, we evaluate LIFE on these two datasets and show 
The accuracy of estimating sparse flows are reported in Tab.~\ref{tab: sparse flow}.
Following RANSAC-Flow~\cite{shen2020ransac}, a correspondence is deemed correct if its endpoint error is less than $\epsilon=1, 3, 5$ pixels.
% RANSAC-Flow filtered correspondences out with a pre-trained segmentation network and RANSAC-based multiple homographies.
% For fair comparison, we use the cycle distance error $d(\mathbf{x})$ as the confidence criterion and individually filters out correspondences whose cycle distance errors are smaller than 0.5~(denoted as LIFE$^*$).
% compare the accuracy separately with and without correspondence filtering. USING WHICH METHOD'S FILTERING PROCESS? YOURS OR THEIRS?
RANSAC-Flow~(R-Flow) regularizes flows by RANSAC-based multiple homographies. 
The images in the MegaDepth are captured at a long distance, such as bird's view images, which can fit the homography model well, so it presents remarkable performance on MegaDepth.
However, this model does not work in general scenes such as the RobotCar, which takes images with car-mounted cameras.
Moreover, we can also use the RANSAC-based multiple homographies as a post-processing strategy to refine flows predicted by other methods if the scenes actually conform to the model.
Compared with other methods that directly predict flows without the homography-based refinement, LIFE outperforms all of them on both RobotCar and MegaDepth.
% LIFE does not achieve a similar large gain on the RobotCar dataset because the RobotCar dataset is much simpler with similar viewpoints.
% SIFT-Flow is a classical handcrafted flow estimation framework, which has very limited performance.
% We also compare LIFE with RANSAC-Flow with a simple correspondence filtering procedure.
% Specifically, our LIFE
% Even without the complex post-processing pipelines used by RANSAC-Flow, LIFE$^*$ largely surpasses RANSAC-Flow$^*$.

\begin{table}[t]
    \centering
    \resizebox{1.0\linewidth}{!}{
    \begin{tabular}{cccc|ccc}
    \hline
    & \multicolumn{3}{c|}{RobotCar} & \multicolumn{3}{c}{MegaDepth}\\
    $\epsilon$ & 1 & 3 & 5 & 1 & 3 & 5\\
    \hline 
    S-Flow~\cite{liu2010sift} & 1.12 & 8.13 & 16.45 & 8.70 & 12.19 & 13.30\\
    DGC~\cite{melekhov2019dgc} & 1.19 & 9.35 & 20.17 & 3.55 & 20.33 & 34.28\\
    GLU~\cite{truong2020glu} & 2.16 & 16.77 & 33.38 & 25.2 & 51.0 & 56.8\\
    GLU-GOC$^+$~\cite{truong2020gocor} & - & - & - & 37.3 & 61.2 & 68.1\\
    R-Flow$^*$~\cite{shen2020ransac}  & 2.10 & 16.07 & 31.66 & $\mathbf{53.47}$ & $\mathbf{83.45}$ & $\mathbf{86.81}$ \\
    \hline
    LIFE & $\mathbf{2.30}$ & $\mathbf{17.40}$ & $\mathbf{34.30}$ & 39.98 & 76.14 & 83.14 \\
    \hline
    \end{tabular}
    }
    \caption{Sparse flow evaluation on RobotCar and MegaDepth. $^*$ denotes that RANSAC-Flow explicitly regularizes flows with RANSAC-based multiple homographies.
    }
    \label{tab: sparse flow}
    \vspace{-3mm}
\end{table}

\begin{table}[t]
    \centering
    \resizebox{1.0\linewidth}{!}{
    \begin{tabular}{cccc|c}
    \hline
    & \multicolumn{3}{c|}{MegaDepth} & HP\\
     & easy &  moderate &  hard & $\epsilon=5$\\
    \hline 
    SIFT~\cite{lowe2004distinctive} & 63.9/25.6 & 36.5/17.0 & 20.8/13.2 & 79.0\\
    SuperPoint~\cite{detone2018superpoint} & 67.2/27.1 & 38.7/18.8 & 24.5/14.1 & 90.5\\
    HardNet~\cite{mishchuk2017working} & 66.3/26.7  & 39.3/18.8 & 22.5/12.3 & - \\
    D2-Net~\cite{Dusmanu2019CVPR} & 61.8/23.6 & 35.2/19.2  & 19.1/12.2 &  - \\
    CAPS+SP kpt.~\cite{wang2020learning} & 72.9/30.5 & 53.5/27.9 & 38.1/19.2 & 90.7 \\
    R2D2~\cite{NEURIPS2019_3198dfd0} & 69.4/30.3  & 48.3/23.9 &  32.6/17.4 & 75.7 \\
    \hline
    LIFE+R2D2 & $\mathbf{78.7}$/$\mathbf{33.1}$  & $\mathbf{62.7}$/$\mathbf{28.7}$ & $\mathbf{45.8}$/$\mathbf{22.4}$ & $\mathbf{91.2}$ \\
    \hline
    \end{tabular}
    }
    \caption{Relative pose estimation on MegaDepth and homography estimation on HPatches. 
    }
    \label{tab: relative pose}
    \vspace{-1mm}
\end{table}

\subsection{Sparse Correspondence Identification.}
% Given a pair of images, we extract keypoints in both images and match them to establish sparse correspondences.

% With the guidance of flow predicted by LIFE, we can establish more and higher quality correspondences from feature matching.
We evaluate LIFE's effectiveness on sparse correspondence identification based on the HPatches dataset.
% With the flow estimated by LIFE, we can alleviate the descriptor ambiguity during feature matching.
% We evaluate the flow-guided feature matching on HPatches to qualitatively illustrate the effectiveness of disambiguity.
Following the protocol in D2-Net~\cite{Dusmanu2019CVPR}, we use the mean matching accuracy~(MMA) and the number of matches as evaluation metrics.
% We compare to several baselines: multi-scale
% D2-Net~\cite{Dusmanu2019CVPR}(D2-Net MS),
% LF-Net~\cite{ono2018lf},
% DELF~\cite{noh2017large},
% Hessian affine detector~\cite{mikolajczyk2004scale} with RootSIFT descriptor~\cite{lowe2004distinctive, arandjelovic2012three}(HesAff),
% SuperPoint \cite{detone2018superpoint},
% SuperPoint detector with CAPS descriptor~\cite{wang2020learning}, as well as R2D2~\cite{NEURIPS2019_3198dfd0}.
% CAPS+SP kpt. is the state-of-the-art, which uses the keypoint detected by SuperPoint and the CAPS descriptor.
As shown in Fig.~\ref{Fig: mma}, LIFE can significantly increase both of the MMA and match number of local features~(LIFE+R2D2 v.s. R2D2 and LIFE+SP v.s. SuperPoint).
% and outperforms other methods when the threshold is larger than 2.
LIFE+R2D2 is slightly inferior at smaller thresholds compared to other methods because the detector of R2D2 is not accurate enough.
In contrast, LIFE+SP shows remarkable performance at all thresholds.
% LIFE+R2D2 is slightly 

\subsection{Whole-image Transformation Estimation}
% To show the practicability of LIFE-based sparse correspondences in downstream tasks, 
In all downstream whole-image transformation estimation tasks,
we can see the significant improvement of LIFE+R2D2 based on R2D2, which demonstrates both of the remarkable flow prediction performance of LIFE and the practicability of LIFE-based sparse correspondence identification.
We evaluate the LIFE+R2D2 on homography estimation, relative pose estimation, and visual localization.

\noindent \textbf{Homography estimation on HPatches.} 
We use the corner correctness metric introduced by SuperPoint \cite{detone2018superpoint}, which transforms the four corners of an image respectively using the estimated homography and the ground truth homography, and compute the average pixel error of the transformed four corners.
An estimated homography is deemed correct if the average pixel error is less than $\epsilon=5$ pixels.
LIFE increases the accuracy of R2D2 by 15.5\% and achieves state-of-the-art performance~(Tab.~\ref{tab: relative pose}).

% \begin{table}[t]
%     \centering
%     % \resizebox{1.0\linewidth}{!}{
%     \begin{tabular}{cccc}
%     \hline
%      & $\epsilon=1$ & $\epsilon=3$ & $\epsilon=5$\\
%     \hline 
%     SIFT~\cite{lowe2004distinctive} & 40.5 & 68.1 & 77.6\\
%     LF-Net~\cite{NEURIPS2018_f5496252} & 34.8 & 62.9 & 73.8\\
%     SuperPoint~\cite{detone2018superpoint} & 37.4 & 73.1 & 82.8\\
%     D2-Net~\cite{Dusmanu2019CVPR} & 16.7 & 61.0 & 75.9\\
%     ContextDesc~\cite{luo2019contextdesc} & 41.0 & 73.1 & 82.2\\
%     R2D2~\cite{NEURIPS2019_3198dfd0} & 40.0 & 75.0 & 84.7\\
%     \hline
%     LIFE & & & \\
%     % LIFE+SIFT &  &  & \\
%     LIFE+R2D2 &  &  & \\
%     \hline
%     \end{tabular}
%     % }
%     \caption{Homography estimation accuracy on HPatches.
%     }
%     \label{tab: homography}
% \end{table}

\noindent \textbf{Relative pose estimation on MegaDepth.}
% We evaluate the relative pose estimation on the MegaDepth dataset and the YFCC100M dataset.
We divide the MegaDepth test set into three difficulty levels according to the ground truth relative rotation angle: easy([0$^\circ$, 15$^\circ$]), moderate~([15$^\circ$, 30$^\circ$]) and hard ([30$^\circ$, 60$^\circ$]).
We report the rotation/translation accuracy in Tab.~\ref{tab: relative pose}.
The relative pose is deemed correct if the angle deviation of its rotation or translation is less than $10^\circ$.
LIFE+R2D2 significantly improves R2D2 and outperforms other methods by large margins.

\noindent \textbf{Visual localization in Aachen.}
We evaluate the visual localization on the challenging Aachen DayNight benchmark~\cite{sattler2018benchmarking}.
The reference images used to build the SfM map are all taken during daytime while the query images are capture at nighttime. 
We report the percentage of query images localized within three given translation and rotation thresholds at nighttime~(Tab.~\ref{tab: visual localization}).
LIFE+R2D2 improves the localization accuracy at all thresholds and ranks 1st on the benchmark at the threshold of (0.5m, $5^\circ$) and (5m, $10^\circ$).

\begin{table}[t]
    \centering
    % \resizebox{1.0\linewidth}{!}{
    \begin{tabular}{cccc}
    \hline
     & 0.25m,$2^\circ$ &  0.5m,$5^\circ$ &  5m,$10^\circ$\\
    \hline 
    SuperPoint~\cite{sarlin2019coarse} & 75.5  & 86.7  & 92.9 \\
    D2-Net~\cite{Dusmanu2019CVPR} & 84.7  & 90.8  & 96.9 \\
    SuperGlue~\cite{sarlin2020superglue} & $\mathbf{86.7}$  & 93.9  & $\mathbf{100.0}$ \\
    R2D2~\cite{NEURIPS2019_3198dfd0} & 80.6  & 90.8  & 96.9 \\
    \hline
    LIFE+R2D2 & 81.6 & $\mathbf{94.9}$ & $\mathbf{100.0}$\\
    \hline
    \end{tabular}
    % }
    \caption{Visual localization on Aachen (night).
    }
    \label{tab: visual localization}
    \vspace{-4mm}
\end{table}

\subsection{Ablation study}

\noindent \textbf{Training loss ablation.}
% We carry out a thorough ablation study to investigate the effectiveness of each proposed module.
We investigate individual components of our LIFE on KITTI and Hpatches with the AEPE metric.
K-12 and K-15 denotes KITTI flow 2012 and KITTI flow 2015. HP-V and HP-I (T) denotes the \textit{Viewpoint} subset and the \textit{Illumination~(trans)} subset in HPatches.
We assess the effectiveness of our methods by sequentially adding the proposed training losses.
T(B'B) denotes that only supervising the network with the flows in one direction of $B'\leftarrow B$ from the synthetic geometric transformation $\mathbf{T}$.
We can see a significant error reduction from T(B'B) to the BiT loss, which indicates the significance of supervising flow prediction in both directions.
The KITTI contains consecutive images that share similar lighting conditions.
Compared with the SED loss, training with the BiT loss produces less errors on the KITTI because it can provide pixel-to-pixel ground truth.
However, it presents worse performance on the HPatches because the synthesized image does not change the illumination and is not in line with the actual situations.
In contrast, the SED loss works with image pairs captured in real scenes so it achieves less error on the HPatches.
We also test imposing cycle consistency on all flows (denoted by ``FC'') and adaptive cycle consistency only on small-error pairs (denoted by ``AC'').
%while AC indicates the adaptive cycle consistency loss as introduced in the method.
FC impacts the flow estimation because occluded regions in the image do not satisfy cycle consistency, and AC can improve SED on viewpoint change cases.
The final model~(SED+BiT+AC) that unifies BiT, SED, and AC in the created image triplets achieves the best performance, which demonstrates the SED loss and the BiT loss mitigates their drawbacks well.

\begin{figure}[t]
    \centering
    \begin{subfigure}[c]{0.7\linewidth}
    \includegraphics[width=\linewidth, trim={0mm 0mm 0mm 0mm}, clip]{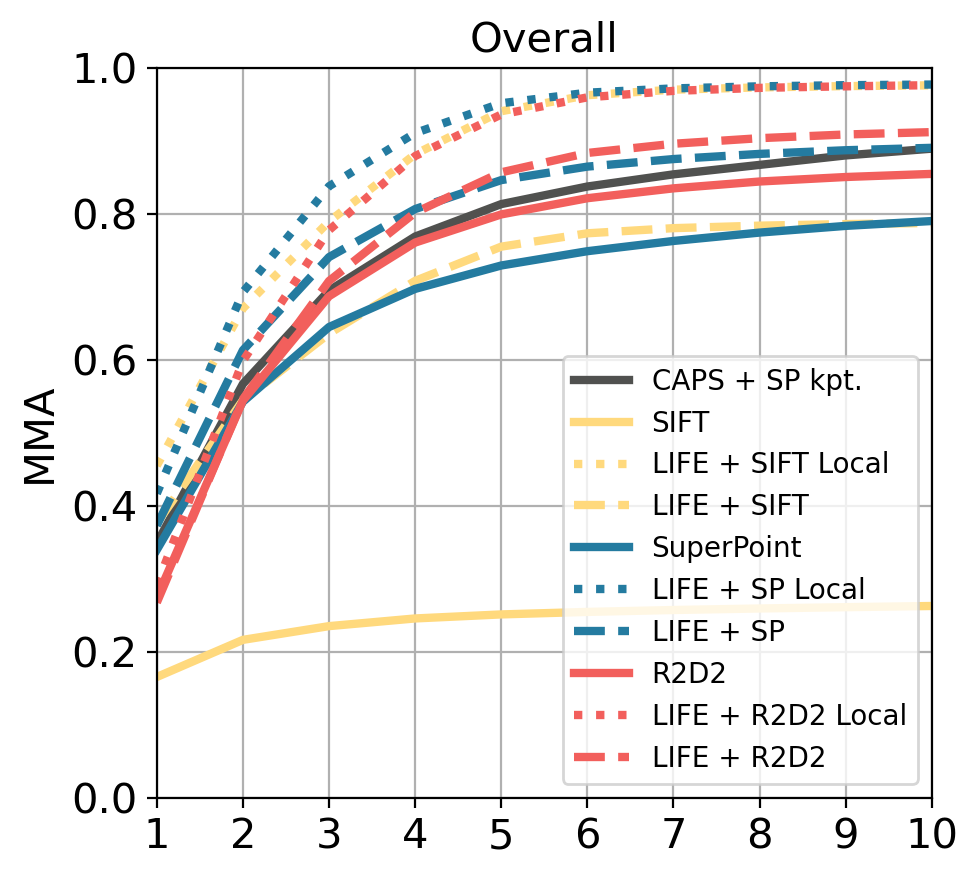}   
    \end{subfigure}
    \begin{subfigure}[c]{0.9\linewidth}
    \centering
    \resizebox{1.0\linewidth}{!}{
    \begin{tabular}{ccccc}
    \hline
     & \#Feat & \multicolumn{3}{c}{\#Matches} \\
     & & raw & LIFE(local) & LIFE(all) \\
    \hline 
    SIFT~\cite{lowe2004distinctive} & 4.4K  & 1.8K & 1.8K & 2.2K\\
    % LIFE+SIFT(l) & 4.4K & \\
    % LIFE+SIFT &  & \\
    SuperPoint~\cite{detone2018superpoint} & 1.7K & 0.9K & 0.9K & 1.0K \\
    % LIFE+SP(l) &  & \\
    % LIFE+SP &  & \\
    R2D2~\cite{NEURIPS2019_3198dfd0} & 5K & 1.8K & 1.9K & 2.1K \\
    % LIFE+R2D2(l) &  & \\
    % LIFE+R2D2 &  & \\
    \hline
    \end{tabular}
    }
    \end{subfigure}

    \caption{LIFE with different local features.
    }
    \label{Fig: ablation}
\end{figure}

\noindent \textbf{LIFE with different local features.}
We test LIFE with different local features~(denoted by ``LIFE+*''), including SIFT, SuperPoint, and R2D2 on HPaches.
As LIFE-based sparse correspondences are computed in two stages, we also report the performance of outcome matches in stage 1~(denoted by ``LIFE+* local''), which are calculated with the guidance of flows predicted by LIFE.  
We report the MMA, feature number, and match number of corresponding methods in Fig.~\ref{Fig: ablation}.
% We set CAPS+SP kpt. as a baseline for comparison.
LIFE consistently improves the MMA and increases the match number.
\textit{LIFE+SIFT local}, \textit{LIFE+SP local}, and \textit{LIFE+R2D2 local} all achieves remarkable MMA scores and remains similar match number, which demonstrates the superior performance of LIFE.
After introducing the matches established in stage 2~(``LIFE+* local'' to ``LIFE+*''),
the match number increases while the MMA scores decreases because the matches calculated in stage 1 are better than 
these supplemented matches.

\begin{table}[t]
    \centering
    \resizebox{0.9\linewidth}{!}{
    \begin{tabular}{c|c|c|c|c}
    % \hline
    Training loss & K-12 &  K-15 &  HP-V & HP-I(T)\\
    \hline 
    T(B'B) & 9.03 & 27.94 & 61.03 & 48.15 \\
    BiT & 3.71 & 11.83 & 16.03 & 37.6\\
    SED & 5.62 & 16.26 & 11.69 & 15.17\\
    SED+FC & 6.48 & 16.83 & 17.79 & 23.64\\
    SED+AC & 4.89  & 14.99 & 12.57 & 15.47 \\
    SED+BiT+AC~(Tri) & 2.59 & 8.30 & 5.90 & 9.19 \\
    \hline
    \end{tabular}
    }  
    \caption{Ablation study of training losses.
    }
    \label{tab: flow ablation}
    \vspace{-3mm}
\end{table}

%% file: 05-conclusion.tex
We have proposed the weakly supervised framework LIFE that learns flow estimation via whole image camera pose transformations, which can predict dense flows between two images with large lighting variations. 
With the assist of LIFE,
we improved the matching of sparse local features, which increased both of the inlier ratio and match number.
The derived sparse correspondences also benefited downstream tasks.
In this paper, the dense flow prediction and sparse feature matching are loosely coupled for sparse correspondences establishment,
which does not maximize the efficiency of information utilization though can directly adopt off-the-shelf sparse features.
Integrating both dense and sparse correspondences estimation into a unified neural network will be the future work.
Moreover, LIFE can predict remarkable flows in challenging scenarios, which may enable more downstream applications.